\documentclass{article}

\usepackage{arxiv}

\usepackage[utf8]{inputenc} 
\usepackage[T1]{fontenc}    
\usepackage{hyperref}       
\usepackage{url}            
\usepackage{booktabs}       
\usepackage{amsfonts}       
\usepackage{nicefrac}       
\usepackage{microtype}      
\usepackage{lipsum}		
\usepackage{graphicx}
\usepackage{natbib}
\usepackage{mathtools}
\usepackage{doi}

\title{Evaluation of Non-Negative Matrix Factorization and n-stage Latent Dirichlet Allocation for Emotion Analysis in Turkish Tweets}

\author{ \href{https://orcid.org/0000-0002-7025-2815}{\includegraphics[scale=0.06]{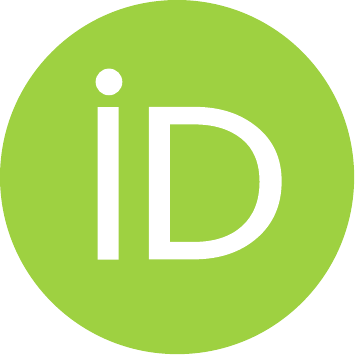}\hspace{1mm}Zekeriya Anil Guven}\thanks{Ege University, Faculty of Engineering, Computer Engineering Department, Izmir, zekeriya.anil.guven@ege.edu.tr. This study is extension version of Comparison Method for Emotion Detection of Twitter Users (http://dx.doi.org/10.1109/ASYU48272.2019.8946435). Please citation this IEEE paper.} \\
	Department of Computer Engineering\\
	Ege University\\
	Izmir, Turkey \\
	\texttt{zekeriya.anil.guven@ege.edu.tr} \\
	\And
	\href{https://orcid.org/0000-0002-4052-0049}{\includegraphics[scale=0.06]{orcid.pdf}\hspace{1mm}Banu Diri} \\
	Department of Computer Engineering\\
	Yildiz Technical University\\
	Istanbul, Turkey \\
	\texttt{diri@yildiz.edu.tr} \\
	\And
	\href{https://orcid.org/0000-0002-4711-7287}{\includegraphics[scale=0.06]{orcid.pdf}\hspace{1mm}Tolgahan Cakaloglu} \\
	Walmart Global Tech\\
	Dallas, USA \\
	\texttt{jackalhan@gmail.com} \\
}

\renewcommand{\shorttitle}{\textit{arXiv} Template}

\hypersetup{
pdftitle={A template for the arxiv style},
pdfsubject={q-bio.NC, q-bio.QM},
pdfauthor={David S.~Hippocampus, Elias D.~Striatum},
pdfkeywords={First keyword, Second keyword, More},
}

\begin{document}
\maketitle

\begin{abstract}
	With the development of technology, the use of social media has become quite common. Analyzing comments on social media in areas such as media and advertising plays an important role today. For this reason, new and traditional natural language processing methods are used to detect the emotion of these shares. In this paper, the Latent Dirichlet Allocation, namely LDA, and Non-Negative Matrix Factorization methods in topic modeling were used to determine which emotion the Turkish tweets posted via Twitter. In addition, the accuracy of a proposed n-level method based on LDA was analyzed. Dataset consists of 5 emotions, namely angry, fear, happy, sad and confused. NMF was the most successful method among all topic modeling methods in this study. Then, the F1-measure of Random Forest, Naive Bayes and Support Vector Machine methods was analyzed by obtaining a file suitable for Weka by using the word weights and class labels of the topics. Among the Weka results, the most successful method was n-stage LDA, and the most successful algorithm was Random Forest.
\end{abstract}


\section{Introduction}
Today, with the continuous increase of data that can be processed over the Internet, there is a need for computer-assisted data analysis. Topic modeling (TM) is one of the effective techniques for text analysis. TM identifies hidden topics based on the words in each document included in the corpus. The determined topics don't depend on the evaluators' perspectives or experiences, it is an objective technique. TM also has some limitations. Loss of interpretability is a major problem of complex learning algorithms such as TM techniques. It is difficult to interpret the issues generated by complex algorithms because outputs with numerical values are produced based on mathematical expressions \citep{Hagen2018}.

TM methods produce interpretable, semantically consistent topics that are represented by listing the most likely words that describe each topic. TM is described as a generative model because it specifies a simple probability procedure that can be generated for the document \citep{Uys2008}. TM identifies documents as a mix of hidden topics. One of the most common algorithms for TM is the Latent Dirichlet Allocation (LDA) algorithm. The LDA algorithm considers all word tokens to be of equal importance, it is a probabilistic model. It is useful for TM, classification, collaborative filtering and information extraction \citep{Martin2015}. In addition, Non-Negative Matrix Factorization (NMF), which is a size reduction and factor analysis method, is also used for TM.

Many machine learning and TM techniques are used to detect the emotion of tweets. In this paper, the most important TM technique, LDA, which is the most important TM technique, and NMF, one of the machine learning methods, were used to detect the emotion of Turkish tweets. In addition, the proposed n-stage LDA algorithm (n-LDA) \citep{Guven2019}, which is language-independent method, was used and these methods were compared with each other. The main contributions of our study to the literature are as follows:
\begin{itemize}
    \item LDA and NMF TM methods are compared for Turkish tweet's emotion analysis.
    \item It is analyzed the effect of proposed n-LDA method. This method aims to delete words with low weights according to the threshold value and give better accuracy with fewer words. 
    \item Experimental results show that n-LDA achieved the best results by using machine learning methods.
\end{itemize}

The content of the paper is as follows. In the second section of this paper, the related works are explained. The methods, the dataset, the preprocessing processes are mentioned in the third section. In the fourth section, the LDA, n-LDA and NMF methods applied on the tweets are analyzed and the results of the methods are shown. In the last section, the evaluation of this study was carried out.

\section{Related Work}
\label{sec:headings}
There are many studies on this subject in the literature. \citet{Suri2017} proposed an approach to extract meaningful data from Twitter. They used LDA and NMF methods to identify topics from data obtained from Twitter. In the results, it was observed that the results of NMF were faster, while the results of LDA were more significant. \citet{Chen2017} compared two interstate word meaning learning models based on NMF and LDA techniques. Both techniques achieved high performance when sufficient number of learning examples were available. They have also proven to be strong against both language and ambiguities. \citet{Belford2018} proposed new measures to assess stability in TM and demonstrated the inherent instability of these approaches. They used LDA for TM and NMF for community learning strategies. Experiments show that a k-fold strategy combining both ensembles and planned baseline significantly reduces indecision and produces more accurate topic models. \citet{Nutakki2014} proposed a hierarchical clustering method called Latent Space (LS) based on LDA and NMF to help organize topics into less fragmented themes. They used a methodology powered by several machine learning techniques, including LDA for TM, hashtag annotations for automated modeling, and TM techniques and NMF for mapping. Thus, they proved that the topics are organized with less fragmented themes and they can relate to each other better. \citet{Lee2012}] proposed a new feature extraction algorithm named dNMF based on GDA and NMF. In addition to the minimum representation error for the standard NMF algorithm, the dNMF algorithm also resulted in higher between-class variance for discriminant power. The improved dNMF algorithm has been applied to emotion recognition for speech, emphasizing emotional differences while reducing the importance of dominant phonetic components. The DNMF algorithm successfully extracted subtle emotional differences and performed much better recognition. \citet{Guven2020} compare the performance of LDA and proposed n-stage LDA with other TM methods. They used Turkish Tweet dataset and Latent Semantic Indexing method showed the best performance compared to other TM methods. \citet{Stevens2012} explored the strengths and weaknesses of each topic model paradigm. They showed that NMF and LDA both learned concise and coherent topics and achieved similar performance on assessments. However, they noted that NMF learned more inconsistent topics than LDA and Singular Value Decomposition. They explained that for applications where the end user will interact with the learned topics, strong consideration should be given to the flexibility and consistency advantages of LDA. \citet{OCallaghan2015} evaluated the consistency and generality of topic descriptions found by the LDA and NMF. Existing and six new corpora were used for this evaluation. They determined a new measure of consistency using a term vector similarity modeled from Word2vec and showed that NMF produced more consistent subjects than LDA. \citet{Habbat2021} performed sentiment analysis on Moroccan tweets. They compared LDA and NMF TM methods for sentiment analysis. Their observed results show that LDA outperforms NMF in terms of their topic coherence. \citet{Chen2019} compared LDA and NMF-based learning schemes for extracting key topics from short texts. Extensive experiments on publicly available short text datasets have shown that NMF tends to produce higher quality topics than LDA.

\section{MATERIAL AND METHODS}
\subsection{Dataset}
The dataset\footnote{https://www.kaggle.com/anil1055/turkish-tweet-dataset} consisting of Turkish tweets was used in this study. In order for the emotion to be understood in tweets, the tweets contain at least one word that expresses emotion (sorry, scared, etc.). The dataset consists of 5 different emotion labels, namely happy (mutlu), sad (üzgün), confused (şaşkın), fear (korku) and angry (kızgın). There are 800 tweets for each emotion. The dataset consists of 2400 tweets for 3 emotion classes and 4000 tweets for 5 emotion classes [17]. 80\% of the dataset was used for training and 20\% for testing.

\subsection{Preprocessing}
In the preprocessing stage, firstly, punctuation marks were separated from the tweets in the dataset. Then, the dataset was converted to lowercase. Since Turkish character conversion can be faulty, letters that are not in English (ö, ş, ı, etc.) are converted to English letters (o, s, i, etc.) with a code. Then, Turkish stopwords were removed from tweets. In addition, a list of words that have no meaning for emotions in the dataset was created. The words in this list have also been deleted from the tweets. Finally, the Zemberek library was used to find the roots of words. Thus, the newly obtained dataset was used for analysis.

\subsection{Latent Dirichlet Allocation}
LDA is a generative probabilistic model for a given text collection. Topics have a probability distribution over words and text documents over topics. Each subject has a probability distribution over the fixed word corpus \citep{Blei2003}. The method exemplifies a mix of these topics for each document. Then, a model is produced by sampling words from this mixture. The model generation process for each document in the text archive is shown as follows \citep{Chen2019}.
\begin{itemize}
	\item $\theta_{n}$ for the nth of documents d in the entire collection of N documents - Dirichlet($\alpha$) selection
	\item For each $W_{n,m}$ word in document d:
	\subitem $Z_{n,m}$ - Multinomial($\theta_{n}$) topic assignment is selected
	\subitem $\phi_Z$ $_{n,m}$ - Dirichlet($\beta$) related topic distribution is found
	\subitem The word $W_{n,m}$ is sampled - Multinomial($\phi_Z$ $_{n,m}$).
\end{itemize}

\begin{figure}[h]
	\centering
	\includegraphics[width=8.0cm]{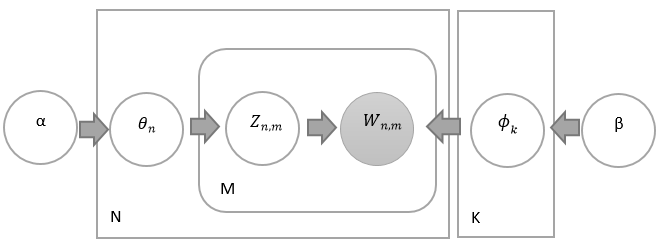}
	\caption{Structure of LDA \citep{Blei2003}}
	\label{fig:fig1}
\end{figure}

The above steps detail the productive process for all N documents in corpus D and for each document. All operations are repeated N times. The process is graphically illustrated in Figure \ref{fig:fig1} ($\alpha$ and $\beta$ are two Dirichlet-priority hyperparameters) \citep{Chen2019}.

The coherence value, which is the TM criterion, is used to determine the number of K topic in the system. The coherence value calculates the closeness of words to each other. The topic value of the highest one among the calculated consistency values is chosen as the topic number of the system \citep{Guven2018a}.

\subsection{n-stage Latent Dirichlet Allocation}
After modeling the system with classical LDA, an LDA-based n-stage method is proposed to increase the success of the model. The value of n in the method may vary according to the size of the data set. With the method, it is aimed to delete the words in the corpus that negatively affect the success. Thus, with the increase in the weight values of the words in the topics formed with the remaining words, the class labels of the topics can be determined more easily.

\begin{figure}[h]
	\centering
	\includegraphics[width=10.0cm]{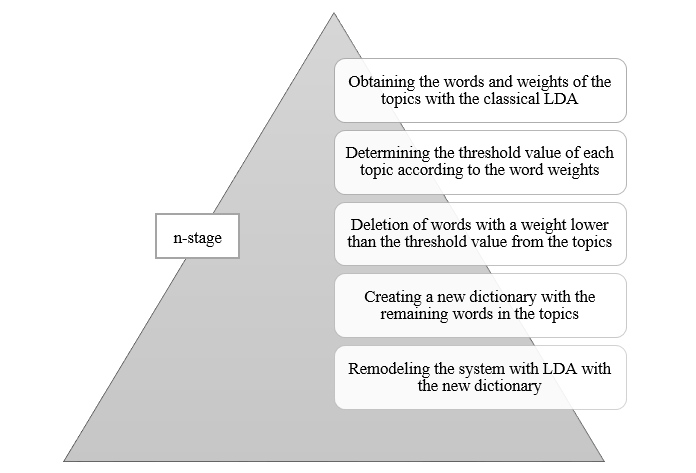}
	\caption{The stages of n-LDA method.}
	\label{fig:fig2}
\end{figure}

The steps of the method are shown in Figure \ref{fig:fig2}. In order to reduce the number of words in the dictionary, the threshold value for each topic is calculated. The threshold value is obtained by dividing the sum of the weights of all the words to the word count in the relevant topic. Words with a weight less than the specified threshold value are deleted from the topics and a new dictionary is created for the model. Finally, the system is re-modeled using the LDA algorithm with the new dictionary. These steps can be repeated n times.

In this method, the reason why we reduce the word count in the dictionary is that the words with low weight in the model cause misclassification. Table \ref{tab:table1} shows the word count with n-LDA for the dataset. It is seen that the word count decreases as the number of stages increases.

\begin{table}[h]
	\caption{The word count according to n-LDA method}
	\centering
    \begin{tabular}{|c|c|}
    \hline
    \textbf{Method} & \textbf{Word count} \\ \hline
    1-LDA              & 2208     \\ \hline
    2-LDA              & 359     \\ \hline
    3-LDA              & 309     \\ \hline
    \end{tabular}
	\label{tab:table1}
\end{table}

The weight values of the remaining words change with n-LDA method. Table \ref{tab:table2} shows the change in weight value of a sample word as the stage progresses. The table shows that the weight value of the word is increasing. Thus, it is foreseen that the class label of the topic will be determined more easily.

\begin{table}[h]
	\caption{The weight value change of the 'yaşa (hurrah in English )' word related to happy}
	\centering
    \begin{tabular}{|c|c|}
    \hline
    \textbf{Method} & \textbf{Weight value} \\ \hline
    1-LDA              & 0.161     \\ \hline
    2-LDA              & 0.366     \\ \hline
    3-LDA              & 0.913     \\ \hline
    \end{tabular}
	\label{tab:table2}
\end{table}

\subsection{Non-Negative Matrix Factorization}
NMF is a size reduction and factor analysis method. Many size reduction techniques deal closely with low-order approximations of matrices. Low-order NMF is special in that it is limited to having only non-negative elements. The non-negative state reflects the natural representation of data in many application areas \citep{DaKuang2015}. Low-order NMF not only allows the user to work with reduced-dimensional models, but also facilitates generally more efficient statistical classification, clustering and organization of data. Thus, it leads to faster searches and queries for patterns or trends \citep{Pauca2006}.

\begin{equation}
	\nu\approx W H
\end{equation}

NMF can be applied to the statistical analysis of multivariate data. Considering a set of multivariate n-dimensional data vectors, the n vectors are placed in the columns of an nXm matrix where the number of samples in the dataset is. This matrix in Equation (1) is then factored into an approximate nXr-dimensional matrix W and an rXm-sized matrix H. Usually r is chosen as less than m. Thus W and H are smaller than the original matrix V. This process results in a compressed version of the original data matrix \citep{Seung2001}.

\section{Experimental Results}
Using the Turkish tweet dataset, the n-LDA model was compared with the classical LDA and NMF. In addition, the data obtained with the system modeled with NMF and LDA were given to machine learning algorithms. F-measure values of Naïve Bayes (NB), Random Forest (RF) and Support Vector Machines (SVM) algorithms were obtained for machine learning algorithms.

For the obtained dataset after the preprocessing steps, 10 coherence values belonging to 3 and 5 labelled classes were calculated. Coherence values for both classes were chosen as the topic number with the highest value among 10 values. The system is modeled with these coherence values. The coherence values and the topic number determined for the 3 and 5 labelled classes for modeling the classical LDA are shown in Table \ref{tab:table3}.

\begin{table}[h]
	\caption{Topic number for each class}
	\centering
    \begin{tabular}{|c|c|c|}
    \hline
    \textbf{Class} & \textbf{Coherence Value} & \textbf{Topic number} \\ \hline
    3              & 0.499                    & 9                     \\ \hline
    5              & 0.484                    & 20                    \\ \hline
    \end{tabular}
	\label{tab:table3}
\end{table}

\begin{figure}[h]
	\centering
	\includegraphics[width=8.0cm]{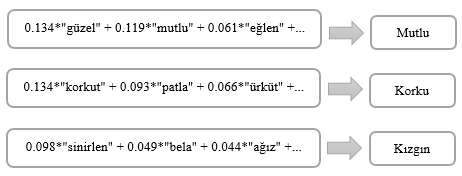}
	\caption{An example for assigning class labels to topics (In English; güzel: beautiful, mutlu: happy, eğlen: amazing, korkut: frighten, patla: burst, ürküt: scare, sinirlen: get angry, bela: darned, etc.).}
	\label{fig:fig3}
\end{figure}

Using the determined topic number for each class, the system is modeled with LDA and the words belonging to the topics and the weights of these words are obtained. By using the words and their weights, the most appropriate class label of the relevant topic is assigned. Figure \ref{fig:fig3} shows an example for determining the most appropriate class label for the topic.

By modeling the system, the topics of the tweets in the dataset were determined. As a result of adding the weights of the words in the tweet separately for all the topics, the topic with the highest value was assigned to the tweet. The success of this system modeled with classical LDA is shown in Table \ref{tab:table4}. As the number of classes increases, the accuracy of the system decreases.

\begin{table}[h]
	\caption{The accuracy of classical LDA}
	\centering
    \begin{tabular}{|c|c|}
    \hline
    \textbf{Class} & \textbf{Accuracy (\%)}  \\ \hline
    3              & 65.8     \\ \hline
    5              & 60.3     \\ \hline
    \end{tabular}
	\label{tab:table4}
\end{table}

In order to analyze its effect on accuracy, n-LDA method was applied to the system. With n-LDA, it is aimed to reduce the number of words in the dictionary and to increase the weight of the words related to emotions. The system was modeled with the two-stage LDA (2-LDA) method with the determined topic numbers. After applying 2-LDA to the model, the 3-stage LDA (3-LDA) method was also applied to the system. For the third stage, the dictionary was created as mentioned before, using the topics in 2-LDA. Thus, the word count in the dictionary has decreased even more compared to 2-LDA. The system is modeled with 3-LDA. The NMF method used in TM was also applied to the system in order to compare with the LDA. For the comparison to be correct, the same topic number was chosen in the NMF method. The words and weights of the subjects extracted with NMF were obtained. The system was modeled by determining the most appropriate class label for the topics. The accuracy values of the 2-LDA, 3-LDA and NMF methods are shown in Table \ref{tab:table5}. The table is show that the accuracy of the LDA method increases as the stage increases and NMF gave better results than n-LDAs.

\begin{table}[h]
	\caption{The accuracy of all methods}
	\centering
    \begin{tabular}{|c|c|c|c|}
    \hline
    \textbf{Methods} & \textbf{3 class} & \textbf{5 class} \\ \hline
    LDA              & 65.8                    & 60.3                     \\ \hline
    2-LDA              & 80.8                    & 70.5                    \\ \hline
    3-LDA              & 81.5                   & 76.4                    \\ \hline
    NMF              & \textbf{89.6}                   & \textbf{82.8}     \\ \hline
    \end{tabular}
	\label{tab:table5}
\end{table}

For each method, a file with a 'csv' extension was created from the weight values of all topics and the class label. The weight value for the topics was calculated by looking at the word weights of the words belonging to each tweet in the topics in the methods. After calculating the weights of the words one by one, the total weight values of the topics of the sentence were obtained. With this file, the accuracy value of all models with machine learning methods was measured. The success of the system were measured for 3 and 5 classes by using 10-fold cross validation with NB, RF and SVM methods in Weka. The accuracy of all methods is shown in Table \ref{tab:table6}. The table shows that as a result of LDAs and NMF methods given to machine learning algorithms, the most successful method was 3-LDA, while the most successful algorithm was RF.

\begin{table}[h!]
	\caption{The accuracy of machine learning methods for all methods}
	\centering
    \begin{tabular}{|l|c|c|c|}
    \hline
    \textbf{} & \textbf{NB} & \textbf{RF} & \textbf{SVM} \\ \hline
    \textbf{NMF - 3 class} & 86.8 & 96.25 & 80.25 \\ \hline
    \textbf{NMF - 5 class} & 71.6 & 92 & 63.1 \\ \hline
    \textbf{Classical LDA - 3 class} & 86.9 & 97.7 & 85.25 \\ \hline
    \textbf{Classical LDA - 5 class} & 72.7 & 93.9 & 73.2 \\ \hline
    \textbf{3-LDA - 3 class} & 86.6 & \textbf{97.8} & 87.6 \\ \hline
    \textbf{3-LDA - 5 class} & 80.8 & \textbf{95.9} & 85.9 \\ \hline
    \end{tabular}
	\label{tab:table6}
\end{table}

\section{CONCLUSION AND DISCUSSION}
In this study, NMF and LDA methods were used to determine the emotion contained in Turkish tweets. In addition to the LDA method, the proposed n-stage method was also analyzed. The n-LDA method used was modeled in 2 and 3 stages. While the 2-LDA method provided an accuracy increase of 10\% to 15\% compared to the classical LDA, the 3-GDA method also provided an accuracy increase of 1\% to 6\% compared to the 2-LDA. The most important reason for these increases is the deletion of words with low weight in the document according to the specified threshold value. Thus, more successful modeling was made with words with better weight. In addition, the success of the csv file created from the weight values and class label for the models in RF, SVM, NB machine learning methods was measured. While the most successful method was RF, the highest accuracy was obtained with the 3-LDA method as 97.8\% for 3-class and 95.9\% for 5-class.

In addition, the success of the model was measured for the NMF method. Compared to the classical and n-LDA, the NMF method was more successful. The accuracy of the NMF method increased between 6\% and 8\% compared to 3-LDA. Then, as with other methods, the accuracy of the NMF method in machine learning methods was measured. The most successful algorithm in NMF has been RF. With the RF method, 96.25\% success was achieved for 3-class and 92\% for 5-class.

\begin{figure}[h]
	\centering
	\includegraphics[width=10.0cm]{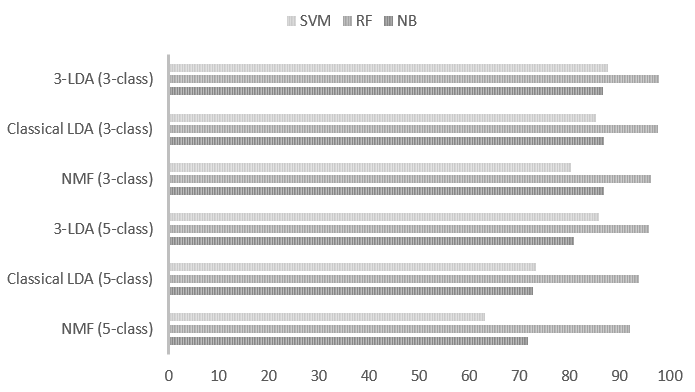}
	\caption{The accuracy value of methods in machine learning algorithms.}
	\label{fig:fig4}
\end{figure}

When NMF, classical LDA and n-LDA methods were compared within themselves, NMF was the most successful method. However, when the accuracy of these methods in machine learning algorithms is compared, the most successful method is 3-LDA. The most successful algorithm for all methods was RF. Figure \ref{fig:fig4} shows the accuracy for machine learning algorithms of all methods. As a result, a positive effect of the n-LDA method on the system was observed. Compared to other methods, it gave more successful results.

In our future works, the n-LDA method; we can use it to determine the type of music and which author the text belongs to, and to compare it with other TM methods. Since the words have a better and higher weighting with the n-LDA method, we think that it will make a positive contribution to the model in which it is tested.

\bibliographystyle{unsrtnat}
\bibliography{references}  
\end{document}